%% file: root.tex
\documentclass[conference]{IEEEtran}

\IEEEoverridecommandlockouts                              

\usepackage{lipsum}
\usepackage{graphicx}
\usepackage{cite}
\usepackage{amsmath,amssymb,amsfonts}
\usepackage{algorithm}
\usepackage{algorithmic}
\usepackage{comment}
\usepackage{caption}
\usepackage{subcaption}
\usepackage{multirow}
\usepackage{float}    
\usepackage{placeins} 
\usepackage{hyperref}
\usepackage{booktabs}
\usepackage{svg}
\input{sections/00_abbreviations}

\title{\LARGE \bf Efficient Cross-Country Data Acquisition Strategy for ADAS via Street-View Imagery}
\author{Yin Wu$^{1, 2}$, Daniel Slieter$^{3}$, Carl Esselborn$^{1}$,\\ Ahmed Abouelazm$^{4}$, Tsung Yuan Tseng$^{1}$, and J. Marius Zöllner$^{2, 4}$
\thanks{$^{1}$Authors are with the CARIAD SE, Germany}
\thanks{$^{2}$Authors are with the Karlsruhe Institute of Technology, Germany}%
\thanks{$^{3}$Authors are with the Esslingen University of Applied Sciences, Germany}%
\thanks{$^{4}$Authors are with the FZI Research Center for Information Technology, Germany}%
}

\begin{document}
\maketitle
\thispagestyle{empty}
\pagestyle{empty}


\begin{abstract}
    \input{sections/0_abstract}

\end{abstract}
\input{sections/1_introduction}
\input{sections/2_related_work}

\input{sections/3_data_collection}

\input{sections/3_Methodology}
\input{sections/4_setting}

\input{sections/5_evaluation}

\input{sections/6_conclusion}

{
    \bibliographystyle{IEEEtran}
    \small
    \bibliography{references}
}

\end{document}

%% file: sections/00_abbreviations.tex
\usepackage[acronym]{glossaries}

\newacronym{adas}{ADAS}{Advanced Driver-Assistance Systems}
\newacronym{ads}{ADS}{Automated Driving Systems}

\newacronym{tsr}{TSR}{Traffic Sign Recognition}
\newacronym{tsd}{TSD}{Traffic Sign Detection}
\newacronym{poi}{POI}{Places of Interest}

\newacronym{knn}{KNN}{k-Nearest-Neighbor}

\newacronym{llm}{LLM}{Large Language Model}
\newacronym{vllm}{vLLM}{Vision Large Language Model}

\newacronym{id}{ID}{In-Distribution}
\newacronym{ood}{OOD}{Out-of-Distribution}

\newacronym{ap}{AP}{Average Precision}
\newacronym{tp}{TP}{True Positive}
\newacronym{tpr}{TPR}{True Positive Rate}
\newacronym{fp}{FP}{False Positive}
\newacronym{fn}{FN}{False Negative}

\newacronym{zod}{ZOD}{Zenseact Open Dataset}

%% file: sections/0_abstract.tex

Deploying \gls{adas} and \gls{ads} across countries remains challenging due to differences in legislation, traffic infrastructure, and visual conventions, which introduce domain shifts that degrade perception performance. Traditional cross-country data collection relies on extensive on-road driving, making it costly and inefficient to identify representative locations. To address this, we propose a street-view-guided data acquisition strategy that leverages publicly available imagery to identify places of interest (POI). Two POI scoring methods are introduced: a KNN-based feature distance approach using a vision foundation model, and a visual-attribution approach using a vision-language model. To enable repeatable evaluation, we adopt a \emph{collect-detect} protocol and construct a co-located dataset by pairing the Zenseact Open Dataset with Mapillary street-view images. Experiments on traffic sign detection, a task particularly sensitive to cross-country variations in sign appearance, show that our approach achieves performance comparable to random sampling while using only half of the target-domain data. We further provide cost estimations for full-country analysis, demonstrating that large-scale street-view processing remains economically feasible. These results highlight the potential of street-view-guided data acquisition for efficient and cost-effective cross-country model adaptation.

%% file: sections/1_introduction.tex
\section{Introduction}
\label{sec:Introduction}

The deployment of \gls{adas} and \gls{ads} in new countries remains challenging due to \emph{domain shift}: models trained in a source market often degrade when exposed to different legislation, traffic infrastructure, and visual conventions~\cite{quinonero2022datasetshift}. As shown in Fig.~\ref{fig:comparison}, even within the European Union, traffic signs may differ across countries (e.g., yellow-background warning signs in Poland vs. white background in Germany), creating non-trivial distribution gaps that affect perception modules.

\input{sections/things/fig_comparison}

A straightforward yet costly solution is to perform large-scale test drives in each target country, which is both resource-intensive and time-consuming. 
Route planning by road type (e.g., urban, highway, suburban) can reduce some effort, but planning those truly informative locations into test drive before visiting there is still infeasible. 
To obtain \emph{a priori} insights before driving, experts increasingly leverage global street-view imagery from platforms such as Google Maps~\cite{googlemaps} and Mapillary~\cite{mapillary_api_docs} to explore candidate locations and identify \glspl{poi} whose visual or semantic attributes differ from those in the source country. 
However, the massive amount of available imagery and its highly variable quality (resolution, blur, viewpoint, and capture device) make both manual inspection and automated \gls{poi} discovery extremely challenging.

Recent advances in vision foundation models~\cite{long2015learning} and \glspl{vllm}~\cite{yang2025qwen3, achiam2023gpt} offer new opportunities for addressing these challenges. 
Such models demonstrate strong zero- and few-shot generalization and can reason about scene semantics directly from visual inputs. In this work, we first systematically propose a cross-country data acquisition strategy, which prioritizes the \gls{poi} for data collection. Then, we leverage the off-the-shelf foundation models and investigate two \gls{poi} scoring method: (1) a \gls{knn}-based feature distance method, which borrow the idea from standard \gls{ood} detection~\cite{sun2022out}. (2) a \gls{vllm}-based visual attributing method that annotates traffic attributes, making \gls{poi} interpretable.

While travelling to different countries and collecting data based on detected \gls{poi} is logistically infeasible, to enable repeatable evaluation, we adopt a \emph{collect-detect} protocol and construct a co-located dataset by pairing the \gls{zod}~\cite{alibeigi2023zenseact} with street-view images retrieved from Mapillary~\cite{mapillary_api_docs}. We study cross-country perception adaptation with Germany as the source domain and Poland, Sweden, and France as target domains, using traffic sign detection as the representative task due to its strong sensitivity to country legislation. Results show that while both methods outperform the random selection baseline, the vLLM-based approach is more stable and achieves comparable effectiveness using only 50\% of the data required by random selection.

In summary, our contributions are:
\begin{itemize}
    \item We introduce a street-view-guided \emph{pre-drive} data acquisition strategy for cross-country model adaptation and propose two \gls{poi} exploration methods based on vision foundation models and \gls{vllm}.
    \item We propose a \emph{collect-detect} evaluation protocol and construct a co-located dataset pairing \gls{zod} with Mapillary street-view imagery, enabling controlled and repeatable assessment.
    \item We conduct comprehensive experiments to demonstrate the effectiveness of our proposed methods. In particular, the \gls{vllm}-based approach achieves comparable performance while requiring only half of the target-domain data used by random selection.
\end{itemize}

%% file: sections/things/fig_comparison.tex
\begin{figure}[htbp]
    \centering
    \includegraphics[width=\linewidth]{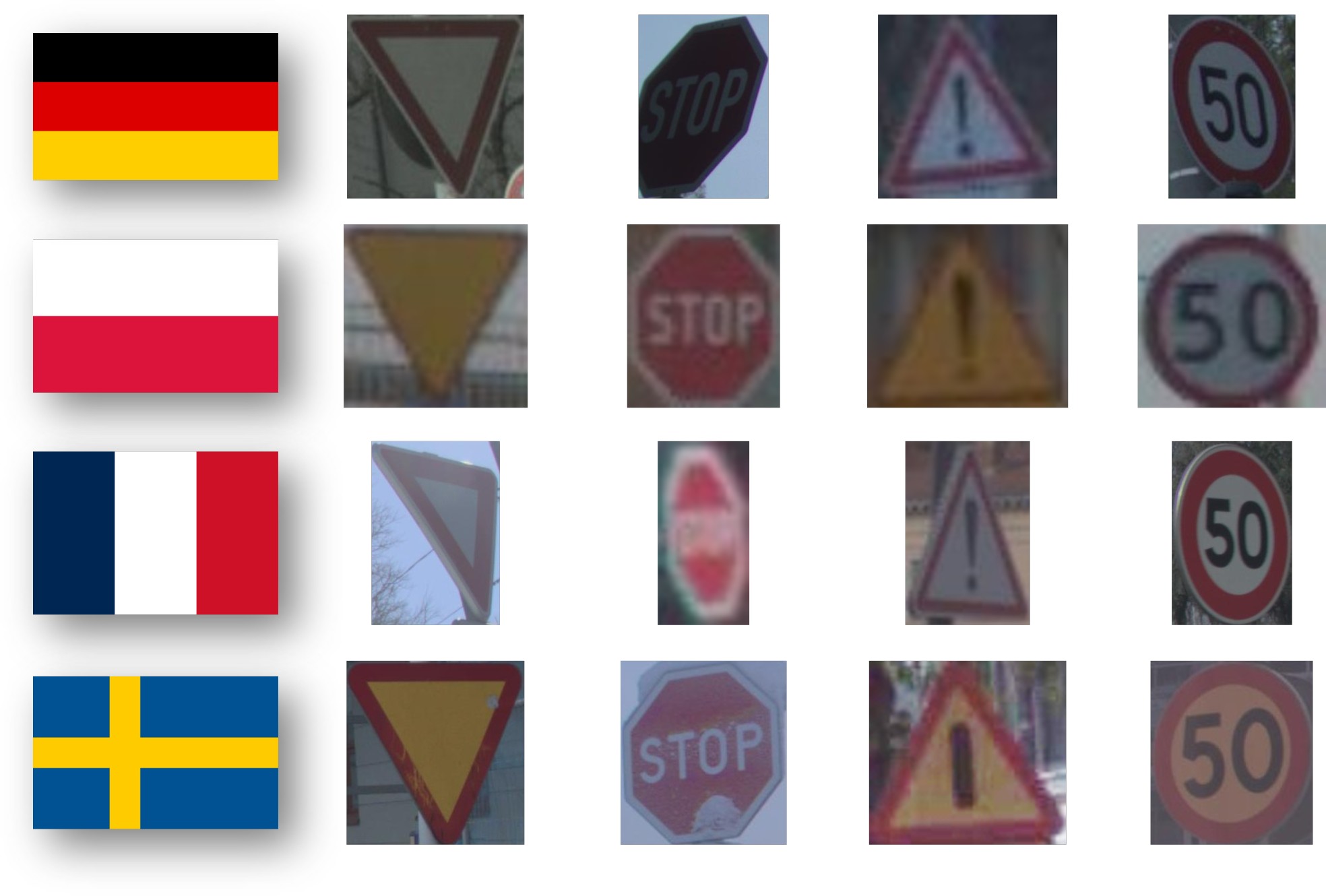}
    \caption{Traffic signs exhibit strong domain shifts across countries. In Poland, warning signs have yellow backgrounds, whereas in Sweden, generic warning signs use a vertical line instead of an exclamation mark~\cite{alibeigi2023zenseact}.}
    \label{fig:comparison}
\end{figure}

%% file: sections/2_related_work.tex
\section{Related Work}
\label{sec:related_work}

In this section, we first discuss \gls{tsd}, which serves as the model under test in our study. 
We then review three closely related research directions: active learning, domain adaptation, and \gls{ood} detection, and explain how they differ from our objective. 
Finally, we summarize recent progress in \gls{vllm} and its applications, which inspires our approach.

\subsection{Traffic Sign Detection and Recognition}

\gls{tsd} and \gls{tsr} are critical perception components in \gls{adas} and \gls{ads}, enabling functions such as intelligent speed assistance (ISA). 
\gls{tsd} can be formulated as a standard object detection problem, where models such as YOLO~\cite{redmon2016you}, Faster R-CNN~\cite{ren2016faster}, and RetinaNet~\cite{lin2017focal} are commonly preferred over transformer-based models~\cite{carion2020end} due to their favorable accuracy-speed trade-offs.  
However, \gls{tsr} presents unique challenges: a large number of fine-grained categories (e.g., different speed limits), highly imbalanced class distributions, and the small physical size of traffic signs within high-resolution driving scenes. 

In this work, we do not aim to develop a new state-of-the-art \gls{tsr} model; instead, we focus on improving cross-country domain adaptation efficiency through data selection and collection strategies. Thus we focus on a binary \gls{tsd} (sign vs.\ background) as a representative \gls{adas} perception task.

\subsection{Active Learning, Domain Adaptation, and Out-of-Distribution Detection}
\label{sec:related_work:active_learning}

Prior studies have extensively discussed the challenges of domain shift and data selection.

Active learning seeks to identify the more informative samples for labeling, thereby reducing annotation costs while maintaining performance~\cite{ren2021survey}.  
Typical approaches exploit model-internal signals, such as uncertainty-based~\cite{beluch2018power} and diversity-based~\cite{gal2017deep} selection, to iteratively query unlabeled samples for annotation.  
Although active learning could, in principle, support our cross-country \gls{poi} selection, standard settings assume that labeled and unlabeled data are drawn from the \emph{same} underlying distribution.  
In contrast, our task involves two distinct resources: high-quality driving data for model training and diverse street-view imagery for \gls{poi} exploration.

Domain adaptation represents another major line of research addressing domain shift between source and target domains~\cite{long2015learning, ganin2016domain,ghifary2016deep, liang2020we}.  
These approaches aim to align feature distributions using metrics such as maximum mean discrepancy~\cite{long2015learning}, adversarial learning~\cite{ganin2016domain}, or reconstruction-based objectives~\cite{ghifary2016deep}.  
Instead of designing a model-centric alignment framework, our work adopts a \emph{data-centric} perspective, exploring how selective data collection can improve finetuning efficiency across countries.

A third related direction is \gls{ood} detection~\cite{sun2022out, guo2017calibration, garg2022leveraging, wu2025laneperf}, which seeks to identify or reject unseen samples during inference to avoid unsafe predictions in safety-critical settings. 
Similar to active learning, many \gls{ood} approaches leverage model-internal uncertainty~\cite{guo2017calibration, garg2022leveraging} or feature-space distance measures~\cite{sun2022out} to detect data that deviates from the training distribution.  
While \gls{ood} detection and our work share the goal of handling distribution shifts, their purposes differ: rather than rejecting \gls{ood} samples, we intentionally identify and leverage them to enhance model generalization and cross-country validation coverage.

Although our problem setting does not exactly fall into any of these paradigms, it is conceptually related to all three.
Inspired by foundation model-guided adaptation~\cite{wu2025laneperf}, we employ pretrained vision encoders and a \gls{knn}-based feature distance mechanism to detect informative locations, functioning as an "active learning style" method.

\subsection{\gls{vllm} and Visual Attribution}

Recent advances in \gls{llm} and \gls{vllm} have demonstrated remarkable zero-shot generalization across diverse perception and reasoning tasks~\cite{achiam2023gpt, yang2025qwen3}. 
Early progress in vision-language alignment can be traced back to CLIP~\cite{radford2021learning}, which performs contrastive learning between paired text and image representations. 
CLIP has shown strong generalization ability across downstream tasks and has since been widely adopted as a pretrained backbone for multimodal understanding~\cite{wu2025laneperf, jiang2023fromcliptodino}.

Building upon these developments, several studies have integrated \gls{llm} and \gls{vllm} into the autonomous driving pipeline. 
For example, Wu \textit{et al.}~\cite{wu2025braking} and Zhao \textit{et al.}~\cite{zhao2024chat2scenario} explore offline scenario extraction and understanding, while Sima \textit{et al.}~\cite{sima2024drivelm} and Tian \textit{et al.}~\cite{tian2024drivevlm} utilize \glspl{vllm} to simulate human-like reasoning in challenging driving situations on a real-time system. Zhong \textit{et al.}~\cite{zhong2025focus} studied visual question answering targeting on small objects.

\input{sections/things/fig_structure}

Within this context, \emph{visual attribution} has emerged as an effective approach for interpretable image classification and pseudo-label generation~\cite{chen2023ovarnet, chen2025hibug2}. 
Following this line of work, we leverage the generalization capability of \gls{vllm} for open-vocabulary traffic sign detection and attribute-based cross-country comparison, enabling a more interpretable understanding of domain-level differences.

%% file: sections/things/fig_structure.tex
\begin{figure*}[htbp]
    \centering
    \includegraphics[width=0.8\linewidth]{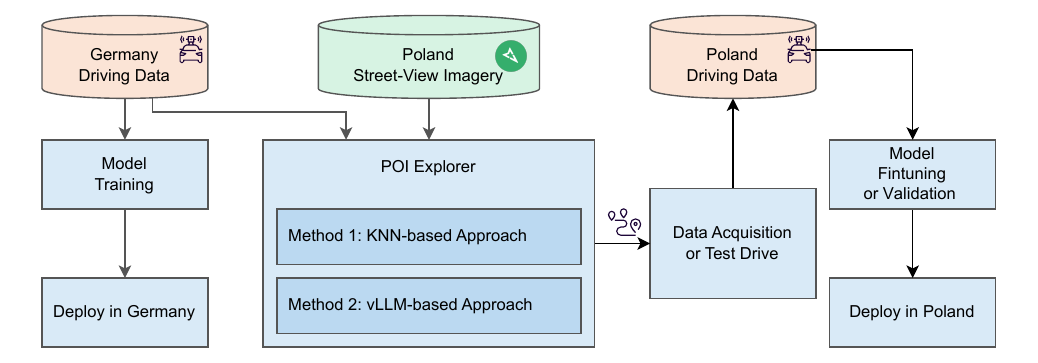}
    \caption{Workflow of the street-view–guided cross-country data acquisition strategy. In this example, Germany serves as the source country providing training data, while Poland is the target deployment country. Instead of using data from the entire target region, the \gls{adas} model is finetuned or validated using only the selected \glspl{poi}, enabling more efficient deployment.}
    \label{fig:structure}
\end{figure*}

%% file: sections/3_data_collection.tex
\section{Problem Statement}
\label{sec:problem_statement}

In this section, we formally introduce the problem of cross-country \gls{poi} exploration. 
We first define the task and its objectives, then describe how to enable controlled and repeatable evaluation through a \emph{collect-detect} protocol. 
Finally, we provide a mathematical formulation of the problem.

\subsection{Problem Protocol}
\label{sec:sec:problem_protocol}

Figure~\ref{fig:structure} illustrates the workflow of our cross-country \gls{poi}-exploration protocol. 
A model is first trained in a source country (e.g., Germany), and the goal is to deploy it in a target country (e.g., Poland). 
Rather than entering the target region without prior knowledge, our strategy leverages publicly available street-view imagery from platforms such as Google Maps and Mapillary to identify locations that exhibit visual or semantic discrepancies relative to the source-domain data.
Locations flagged as \glspl{poi}, those likely to be unfamiliar or challenging for the model, are added to a prioritized test route. 
Test engineers can then perform targeted data collection or function validation during a subsequent test drive in the target country. 
This protocol provides actionable \emph{a priori} insights and reduces unnecessary on-road exploration by focusing attention on locations that matter for cross-country adaptation.

In principle, a more faithful evaluation paradigm for this task is a \emph{detect--collect} protocol: first exploring the target country’s street-view imagery to detect \glspl{poi}, and performing real-world driving to these locations for data collection or validation. Then we finetune the model on the selected data. A larger performance improvement indicates that more discrepant data are selected. 
However, executing such a protocol is logistically infeasible in research settings, as it would require repeated international test drives for every experiment, resulting in prohibitive cost and
time.

To enable systematic evaluation, we adopt a \emph{collect--detect} protocol as a practical and scalable proxy. We start from a pre-collected multi-country driving dataset and retrieve street-view images at the same GPS coordinates of each image in the driving dataset. \gls{poi}-exploration
methods operate on the street-view set and select the top-k discrepant places. The co-located driving data is selected for finetuning the model.

It is important to note that \emph{collect--detect} is not intended to perfectly replicate \emph{detect--collect}. Instead, it provides a controlled, repeatable, and fair environment for comparing different \gls{poi}-exploration strategies under identical data budgets. The goal is to study the \emph{relative} effectiveness of selection methods, rather than to simulate full-scale deployment behavior.

\subsection{Problem Formulation}
\label{sec:sec:problem formulation}

We formulate the previous introduced problem mathematically. Let \(D_{\text{train}}^S\) denote the source-domain training set, and \(D_{\text{train}}^T\) and \(D_{\text{val}}^T\) the training and validation sets from the target domain. The goal of \gls{poi} exploration is to select a subset \(D_{\text{train}}^{T,k} \subseteq D_{\text{train}}^{T}\) of size \(k\) such that finetuning on this subset maximizes target-domain performance. Let \(\mathcal{M}\) be a model pretrained on \(D_{\text{train}}^S\) and \(\mathcal{M}(D_{\text{train}}^{T,k})\) the model after finetuning on the selected subset. With \(\mathcal{P}(\cdot, D_{\text{val}}^T)\) denoting an evaluation function, the data-selection objective is:
\[
D_{\text{train}}^{T,k*}
= \arg\max_{D_{\text{train}}^{T,k} \subseteq D_{\text{train}}^T}
\mathcal{P}\!\left(\mathcal{M}(D_{\text{train}}^{T,k}),\, D_{\text{val}}^T\right).
\]


Since driving data from the target domain is typically unavailable prior to collection, we assume access only to a set of street-view images \(I_T\), grouped by GPS location. Each location \(\ell\) is associated with a set of retrieved street-view images \(I(\ell) \subseteq I_T\). A \gls{poi}-exploration method computes an image-level discrepancy score for each \(x \in I(\ell)\), and aggregates them into a location-level score \(s(\ell)\). The top-\(k\) locations are then selected:
\[
L_T^{k} = \operatorname{TopK}_{\ell}(s(\ell)).
\]

The corresponding driving data at these locations form the fine-tuning subset:
\[
D_{\text{train}}^{T,k} = \Phi(L_T^{k}),
\]
where \(\Phi\) maps selected locations to their associated driving logs. Thus, \gls{poi} exploration reduces to designing a scoring function \(s : \ell \rightarrow \mathbb{R}\) whose induced selection \(D_{\text{train}}^{T,k}\) yields maximal downstream performance in the target domain.

%% file: sections/3_Methodology.tex
\input{sections/things/fig_two_methods}

\section{Methodology}
\label{sec:methodology}


In the following two sections, we introduce our proposed two discrepancy scoring methods: \gls{knn}-based feature distance and \gls{vllm}-based visual attributing. As shown in Figure~\ref{fig:two_methods}, both methods first extract features / attributes on the source driving data and target street-view data. Then, using a features / attributes distance to calculate the score.

\subsection{Method 1: KNN-based Feature Distance Scoring}
\label{sec:method1}



As discussed in Section~\ref{sec:related_work:active_learning}, our problem does not follow a standard active-learning setting. To quantify cross-domain differences, we adopt a \gls{knn}-based feature distance scoring method inspired by~\cite{sun2022out}. Instead of the image encoder of the \gls{adas} perception model, we extract a representation \(\mathbf{f}(x)\) using a pretrained foundation vision encoder. Source-domain features are obtained from driving images, whereas target-domain features come from street-view imagery. Given the substantial variability in street-view data (e.g., viewpoints, lighting, and capture devices), a foundation model provides more stable and domain-agnostic representations than the task-specific encoder.




\input{sections/things/fig_prompt_text}

Let $\mathcal{F}_S=\{\mathbf{f}(x_i^S)\}$ denote the feature set of source-domain images. For a target location (driving log) $l$, multiple street-view images may be retrieved. We compute a per-image discrepancy score based on the $k$-th nearest neighbor distance to the source-domain feature vectors:
\[
s(x)=d_k\big(\mathbf{f}(x),\mathcal{F}_S\big),
\]
where $d_k$ is the $k$-th smallest Euclidean distance in the normalized feature space. 

To obtain a location-level score $s(l)$, we aggregate scores across all images associated with the same driving location. We take the minimum discrepancy to mitigate viewpoint variability:
\[
s(l)=
\begin{cases}
\min\limits_{x\in I(l)} s(x), & \text{if } I(l)\neq\emptyset,\\[6pt]
0, & \text{if no street-view image is available},
\end{cases}
\]
where $I(l)$ is the set of retrieved images for location $l$. Locations without image returns are assigned a score of zero.

\subsection{Method 2: \gls{vllm}-based Attribute-Level Scoring}
\label{sec:method2}

While the \gls{knn}-based method captures global visual deviations, our second scoring strategy focuses on fine-grained, traffic-sign-specific semantics. Instead of comparing image-level embeddings, we rely on a \gls{vllm} to extract structured traffic-sign attributes from street-view imagery, enabling more interpretable and controllable \gls{poi} identification.

\paragraph*{Attribute extraction.}
As shown in Figure~\ref{fig:prompt}, each street-view image is processed with a structured prompt to detect all visible traffic signs and annotate them using a predefined attribute ontology. Each detected sign is represented as an attribute vector
\[
\mathbf{a} = \big(
a_{\text{category}},\, a_{\text{shape}},\, a_{\text{color\_border}},\, 
a_{\text{color\_background}},\,
\]
\[
\quad a_{\text{color\_symbol}},\, a_{\text{symbol}},\, 
a_{\text{text}},\, a_{\text{language}}
\big),
\]

where each $a$ corresponds to one traffic-sign attribute. A driving location $l$ may have multiple retrieved images, and each image may contain multiple signs. We aggregate all detected signs for location $l$ and remove duplicates, yielding a unique sign-attribute set
\[
A(l) = \{\mathbf{a}_1, \mathbf{a}_2, \dots, \mathbf{a}_M\}.
\]

To calculate the discrepancy score from attributes, we first construct the set of all unique sign attribute vectors based on the driving data of the source country:
\[
A_S = \{\mathbf{a}_i^S\}.
\]
For a target-domain sign $\mathbf{a} \in A(l)$, we measure its semantic deviation from the source by comparing it to the most similar source sign in $A_S$ using a Hamming distance over attributes:
\[
d(\mathbf{a}, \mathbf{a}') = \sum_{j=1}^{K} \mathbf{1}\!\left[a_j \neq a'_j\right],
\]
where $K$ is the total number of attributes to describe a traffic sign ($K=8$ in our setup),
and
\[
d(\mathbf{a}, A_S) = \min_{\mathbf{a}' \in A_S} d(\mathbf{a}, \mathbf{a}').
\]
The location-level discrepancy score is then defined as the sum of per-sign deviations:
\[
s(l) = \sum_{\mathbf{a} \in A(l)} d(\mathbf{a}, A_S).
\]
A higher score indicates that the location contains more traffic signs whose attributes differ from those observed in the source domain (e.g., new shapes, color schemes, symbols, or languages). Such locations are prioritized as discrepant \gls{poi}.

%% file: sections/things/fig_two_methods.tex
\begin{figure}[t]
\centering
\setlength{\tabcolsep}{2pt}
    \includegraphics[width=\linewidth]{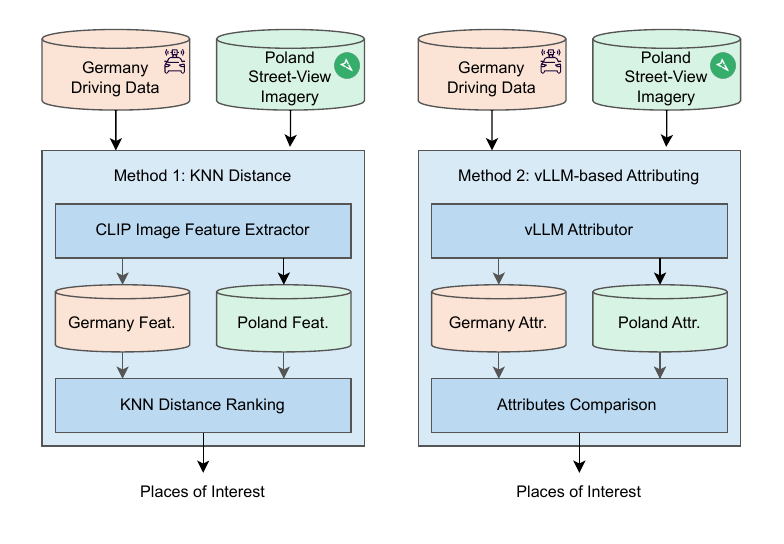}
\caption{Two proposed foundation model-based methods to detect \gls{poi}.}
\label{fig:two_methods}
\end{figure}

%% file: sections/things/fig_prompt_text.tex
\begin{figure}[t]
\centering
\tiny
\ttfamily
\begin{minipage}{0.95\linewidth}
\begin{verbatim}
You are given an image. Detect all traffic signs and return
structured JSON data.

For each traffic sign, include:
- a descriptive name (free text)
- structured attributes
- high-level category (from a fixed set)

Use the format:
[
  {
    "name": "<short_name_with_underscore>",
    "category": "<one of: regulatory, warning, informational,
                 guide, construction, school_zone, other>",
    "attributes": {
      "shape": "<one of: circle, triangle, rectangle, square,
                 octagon, diamond, inverted_triangle>",
      "color": {
        "border": "<one of: red, white, yellow, black,
                   blue, green, none>",
        "background": "<one of: red, white, yellow, black,
                       blue, green, none>",
        "symbol": "<one of: red, white, yellow, black,
                   blue, green, none>"
      }
    },
    "symbol": "<one of: arrow, pedestrian, car, bicycle,
                stop_hand, number, none, other>",
    "text": "<extracted text or 'none'>",
    "language": "e.g., English, German, Chinese, etc."
  }
]

Rules:
- Only return actual traffic signs
- If the sign has no text, use "text": "none"
- If the sign uses only symbols (no text), set "language": "none"
- If no signs are detected, return []
- Do not return signs on road
\end{verbatim}
\end{minipage}
\caption{Prompt used for traffic sign attribute extraction.}
\label{fig:prompt}
\end{figure}

%% file: sections/4_setting.tex
\input{sections/things/tab_data_statistic}
\input{sections/things/fig_example_images}

\section{Experimental Setup}
\label{sec:experimental_setup}


\subsection{Dataset Preparation}

Following the \emph{collect--detect} protocol introduced in Section~\ref{sec:sec:problem_protocol}, we use the \gls{zod} dataset~\cite{alibeigi2023zenseact} as the driving-data source. \gls{zod} provides multi-country driving logs with high-precision GPS coordinates and traffic-sign annotations. We focus on the four countries with the largest data volume: Germany (DE), Poland (PL), Sweden (SE), and France (FR). Germany serves as the source domain for pretraining, while the other three serve as target domains. Each country-specific subset is split into training and validation sets using a 9:1 ratio.

Street-view imagery is obtained using the Mapillary API~\cite{mapillary_api_docs}, which supports GPS-based retrieval. For each driving-log coordinate in the training split, we request all available street-view images within a 10\,m radius. The number of returned images varies across locations (from zero to multiple). Up to 10 images are retained per location, ranked by proximity to the query coordinate. The final retrieval statistics are summarized in Table~\ref{tab:data_statistic}: overall, more than 80\% of driving-log locations successfully return at least one street-view image. 
For locations where no street-view imagery is available, we assign a default POI score of zero. 
Figure~\ref{fig:example_images} illustrates typical retrieved samples, where the images originate from different viewpoints and capture devices, highlighting the challenges posed by street-view data.

\subsection{Model Under Test}

We choose \gls{tsd} as the \gls{adas} perception model to do cross-country deployment for three reasons: (1) \gls{tsd} is essential for traffic-rule interpretation, (2) detection performance is highly sensitive to country-specific regulatory differences, and (3) high-quality annotations are already provided in \gls{zod}.

We adopt RetinaNet~\cite{lin2017focal} implemented in Detectron2~\cite{wu2019detectron2}, owing to its favorable accuracy-efficiency trade-off for on-board deployment. The model is first pretrained on the Germany training set for 20 epochs with $batch\_size=8$ using the Adam optimizer with a learning rate of 0.0025. For adaptation, the model is finetuned using \(k\) samples from the target-domain training set, where the selected \(k\)-subset is determined by a \gls{poi} exploration method.

Performance is evaluated using \gls{ap} at an IoU threshold of 0.5. Higher \gls{ap} improvements after finetuning indicate that the selected subset is more informative. We evaluate multiple data budgets, expressed as percentages of the target-domain training set. Let \(\alpha \in \{5\%, 10\%, 15\%, 20\%, 25\%, 50\%, 75\%\}\) denote the sampling ratio. For each \(\alpha\), the actual subset size is \(k = \alpha \cdot |D_{\text{train}}^T|\), and we finetune the pretrained model on \(D_{\text{train}}^{T,k}\). Each finetuning run spans 10 epochs with the same hyperparameters as pretraining.

\subsection{POI-Exploration Method Details}

We compare three \gls{poi} exploration methods:

\paragraph{Random Selection (Baseline).}
A uniform random subset of size \(k\) is selected without using street-view imagery.

\paragraph{\gls{knn}-based method.}
We use the ViT-L/14 image encoder from CLIP~\cite{radford2021learning} as the image encoder due to its strong generalization ability. For \gls{knn} algorithm, we select $k=10$, following one setup from the original paper~\cite{sun2022out}

\input{sections/things/fig_eval_num_collection}

\paragraph{\gls{vllm}-based method.}
We employ \texttt{gpt-5-mini} (model version 2025-08-07) from OpenAI~\cite{achiam2023gpt} conduct visual attributing. While single-image API calls require approximately 5\,s per image, we leverage batch-mode processing (up to 50k images per batch), which significantly accelerates scoring and enables large-scale evaluation.

%% file: sections/things/tab_data_statistic.tex
\begin{table*}[t]
\centering
\caption{Statistics of the \gls{zod} and street-view image retrieval results from Mapillary. Over 80\% of logs yielding at least one image. Only data from Poland (PL), Germany (DE), Sweden (SE), and France (FR) are considered, as the remaining countries in \gls{zod} do not contain sufficient samples for country-specific training.}
\label{tab:data_statistic}
\begin{tabular}{lccccc}
\toprule
\textbf{Country} & \textbf{Logs in \gls{zod}}& \textbf{Logs w. Traffic Sign} & \textbf{Has Street-View Image Return} & \textbf{Ratio of Logs with Mapillary Returns}\\
\midrule
PL & 30,483 & 23,207 & 20,022 & 86.28\%\\
DE & 24,816 & 19,492 & 17,094 & 87.70\%\\
SE & 17,647 & 8,280 & 6,912 & 83.48\%\\
FR & 11,785 & 7,666 & 6,560 & 85.57\%\\
\bottomrule
\end{tabular}
\end{table*}

%% file: sections/things/fig_example_images.tex
\begin{figure}[t]
\centering
\setlength{\tabcolsep}{2pt}
\begin{tabular}{cc}
    \subcaptionbox{Image from \gls{zod}\label{fig:example_1}}{%
        \includegraphics[width=0.45\linewidth]{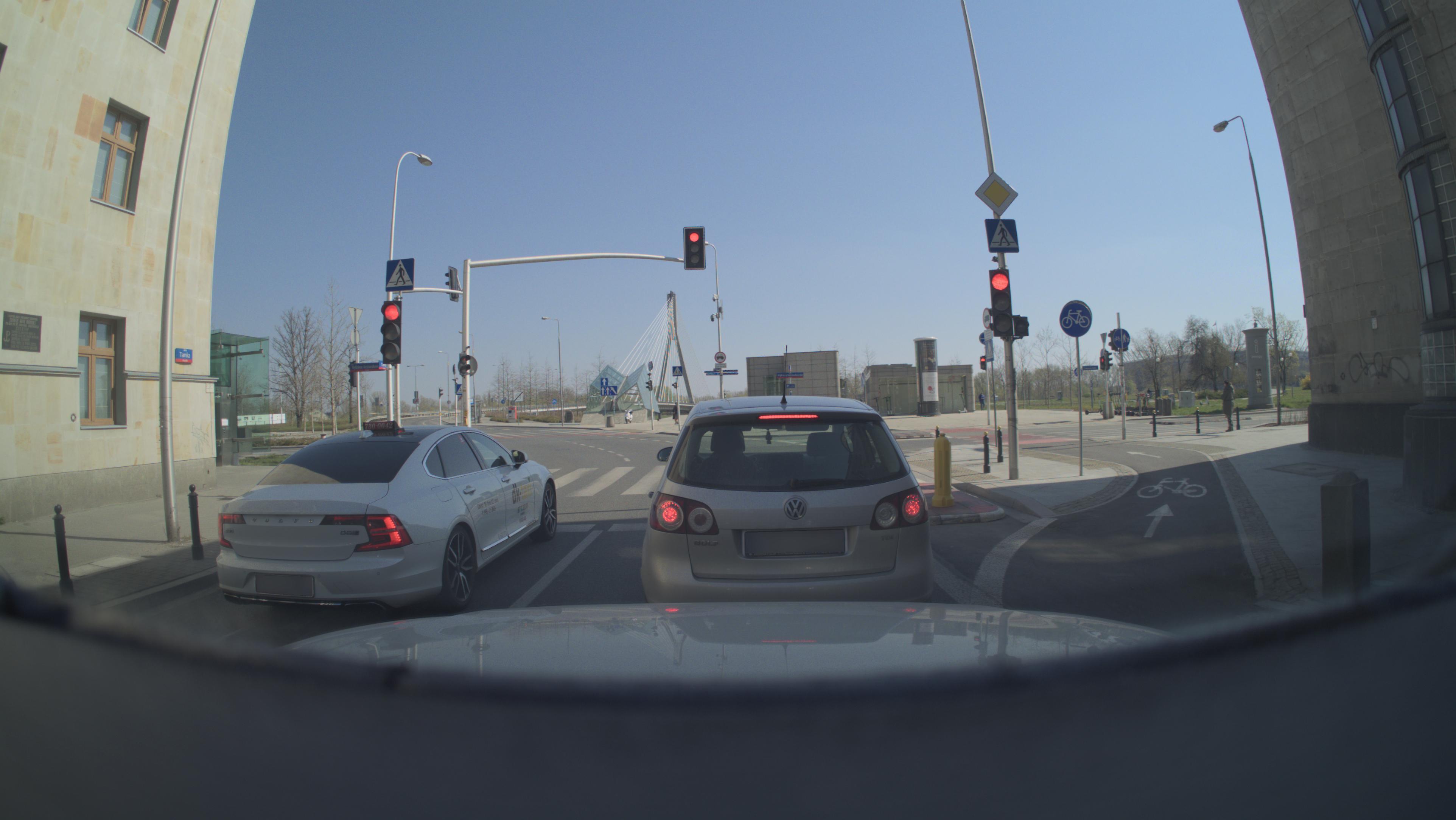}} &
    \subcaptionbox{Pedestrian View\label{fig:example_2}}{%
        \includegraphics[width=0.45\linewidth]{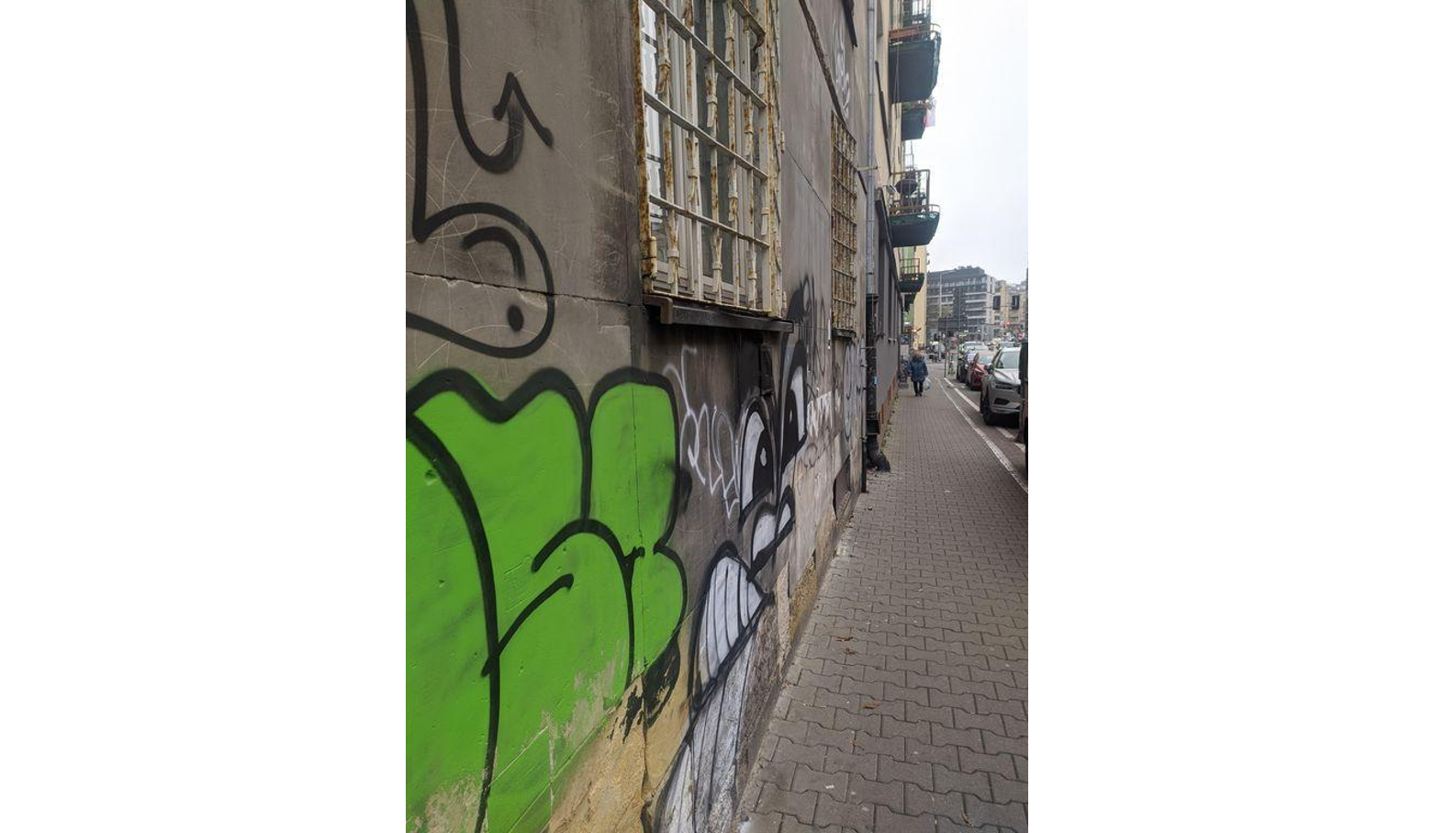}} \\[4pt]
    \subcaptionbox{Bicycle View\label{fig:example_3}}{%
        \includegraphics[width=0.45\linewidth]{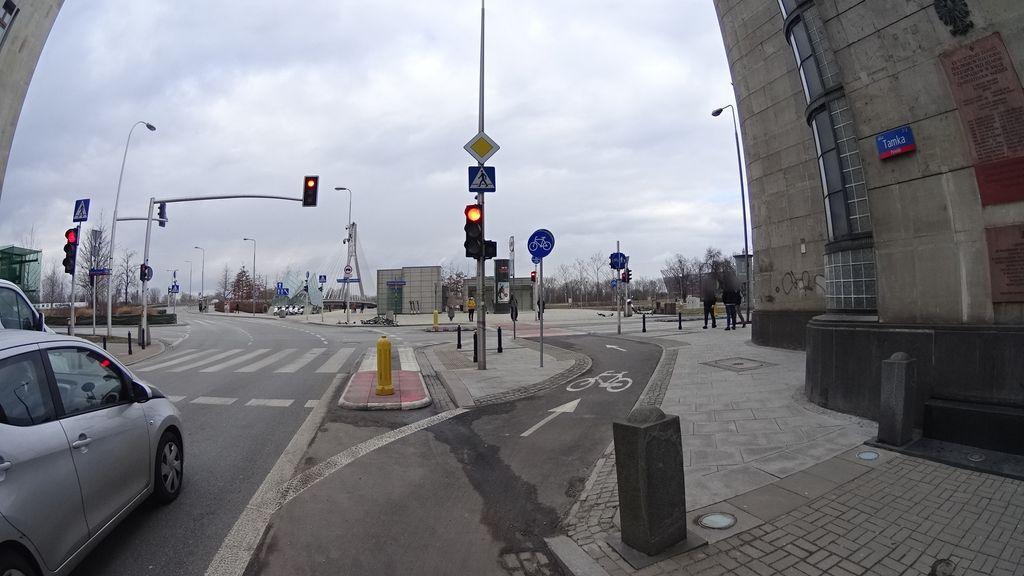}} &
    \subcaptionbox{Vehicle View\label{fig:example_4}}{%
        \includegraphics[width=0.45\linewidth]{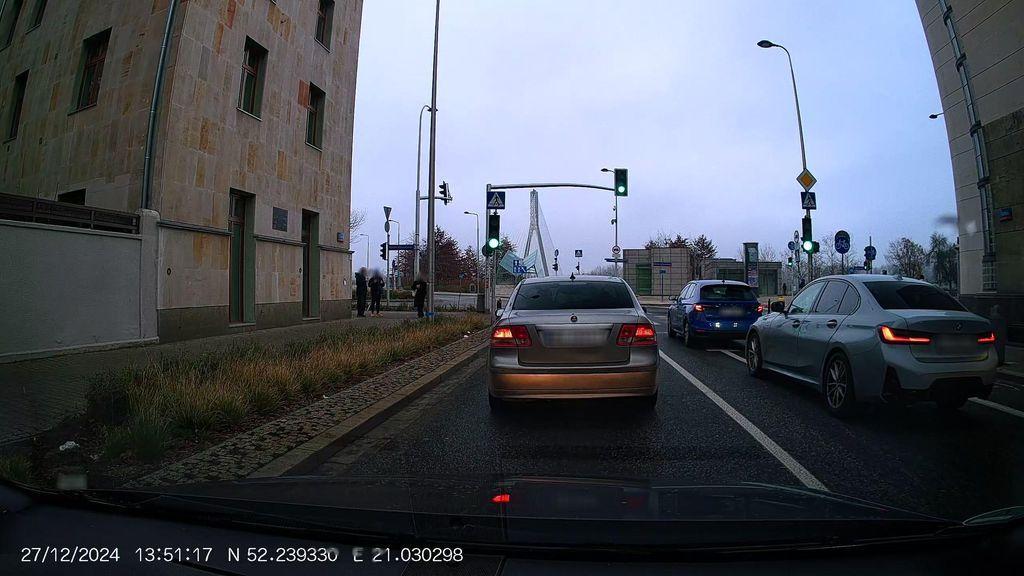}}
\end{tabular}
\caption{Examples of co-located images. (a) shows the on-board image from the \gls{zod}~\cite{alibeigi2023zenseact} dataset, while (b)-(d) are street-view images within a 10\,m radius from Mapillary~\cite{mapillary_api_docs}.}
\label{fig:example_images}
\end{figure}

%% file: sections/things/fig_eval_num_collection.tex
\begin{figure}[htbp]
    \centering
    \includegraphics[width=\linewidth]{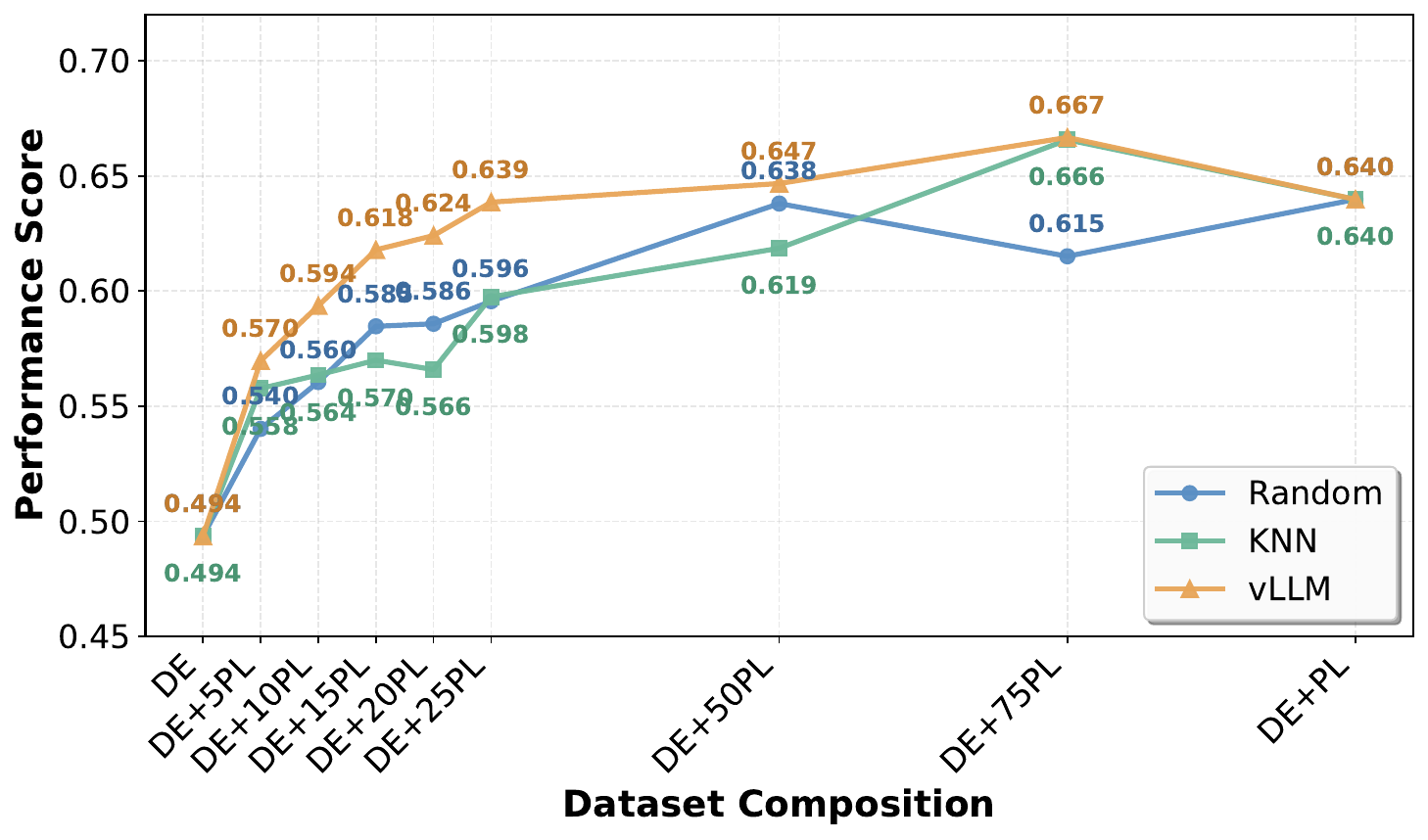}
    \caption{The \gls{tsd} model is pretrained on ZOD Germany dataset and then finetuned with different portion of ZOD (Poland) dataset (5\%, 10\%, 15\%, 20\% 25\%, 50\%, 75\%, 100\%). Three \gls{poi} exploration methods are compared: Random Selection, \gls{knn}-based method, \gls{vllm}-based method.}
    \label{fig:eval_num_collection}
\end{figure}

%% file: sections/5_evaluation.tex
\section{Evaluation}
\label{sec:evaluation}

We evaluate the effectiveness of the street-view-guided data acquisition strategy and the proposed \gls{poi} scoring methods.
Three research questions (RQ) are answered in the following sections:
\begin{itemize}
    \item RQ1: Can proposed methods achieve comparable performance with requiring \emph{less} target-domain data, or equivalently, do they achieve \emph{greater} performance improvement under the same data budget? 
    \item RQ2: Can the strategy generalize across different target countries, demonstrating consistent benefits beyond a single deployment scenario? 
    \item RQ3: Do the selected \glspl{poi} reflect meaningful cross-country differences that can be qualitatively interpreted from street-view imagery?
\end{itemize}


\subsection{RQ1: Finetuning with Different Portions of Target Data}
\label{sec:evaluation_portion}

We first report the finetuning performance when varying the amount of Polish training data added on top of the German pretrained model. Several observations emerge.

\input{sections/things/fig_eval_countries}

As shown in Figure~\ref{fig:eval_num_collection}, finetuning with only a small portion of PL data already yields significant performance gains. For example, using 10--25\% of the target-domain data results in \gls{ap} values of 0.594--0.639 with the \gls{vllm}-based scoring method, approaching the performance of full-data finetuning (0.640). This confirms that exhaustive target-domain data collection is not strictly necessary: a carefully selected subset can recover most of the achievable adaptation benefit.

Second, the proposed \gls{vllm}-based scoring consistently outperforms both random sampling and the \gls{knn}-based method across nearly all data budgets. With only 25\% of the target-domain data, the \gls{vllm} approach already surpasses the performance achieved by using 50\% of the data under random selection. The improvement is particularly pronounced at low data budgets (5--25\%), where selecting semantically distinctive locations is critical.

In contrast, the \gls{knn}-based method provides only a marginal average improvement over random sampling (+0.03 \gls{ap}) and exhibits high variance across runs. This instability arises because CLIP embeddings capture global scene semantics rather than traffic-sign-specific cues, making the feature-distance measure sensitive to irrelevant visual variations in street-view imagery.

Finally, we observe that selectively chosen data can even surpass full-data training. At 75\% of the PL data, \gls{vllm} and \gls{knn} achieve an \gls{ap} of 0.667/0.666, exceeding the full-data finetuning 0.640. According to~\cite{paul2021deep}, this can be explained by the presence of redundancy and noise in the full dataset, and further emphasizes that the goal of \gls{poi} exploration is not to gather more data, but to gather \emph{better} data.




\subsection{RQ2: Finetuning Across Different Countries}
\label{sec:eval_cross_country}
We further evaluate whether the proposed \gls{poi} exploration strategy generalizes across different target countries. Based on the results in Figure~\ref{fig:eval_num_collection}, we adopt a 25\% data budget for all target domains, as performance largely saturates beyond this point. Starting from the German-pretrained model, we fine-tune on 25\% of the training data from Poland, Sweden, and France and report the corresponding validation performance. For reference, we also include the German validation \gls{ap} to indicate in-domain performance. The results are shown in Figure~\ref{fig:eval_countries}.

Across all three target countries, the no-finetuning model performs worse than on Germany, confirming the presence of significant cross-country domain shift and the necessity of target-domain data acquisition. Finetuning with 25\% of the target data yields clear improvements in all cases, closing much of the performance gap. 

Among all data-selection strategies, the proposed \gls{vllm}-based method consistently achieves the highest performance in every country. The gains over random selection are substantial: +0.043 AP in Poland, +0.054 in Sweden, and +0.076 in France. \gls{knn}-based method also improves over random in Sweden and France but exhibits marginal benefit in Poland (+0.002), showing lower stability across domains. These results demonstrate that \gls{vllm}-based visual attribution is more reliable in identifying discrepant \glspl{poi} that support effective cross-country adaptation.

\input{sections/things/fig_eval_exps_v2}

\subsection{RQ3: Qualitative Output Analysis and Failure Cases}
\label{sec:qualitative}

Figure~\ref{fig:eval_exps} shows representative examples from the two methods. Subfigures (a)-(d) come from the top-ranked outputs of the \gls{vllm}-based method. Examples (a) and (b) illustrate successful identification of country-specific traffic signs in Poland. In (c), the sign detected in the street-view image is not visible in the driving data due to a mismatch of viewpoint. Example (d) shows a temporal discrepancy: the guide sign appears in the street-view image, but had been removed by the time the test vehicle passed the location. Under the adopted \emph{collect-detect} protocol, such spatial and temporal mismatches are expected and do not indicate method failure; instead, they reflect real-world factors that a full \emph{detect-collect} pipeline would naturally address.

Figure (e) shows a typical failure from the \gls{knn}-based method. Although with a high discrepancy score, the street-view image does not contain any traffic-scene content. This highlights the limitation of general image embedding for task-specific data acquisition and the advantage of \gls{vllm}-based reasoning in focusing on semantically relevant \glspl{poi}.

\subsection{Cost Analysis}
\label{sec:cost_analysis}
We estimate the computational cost of performing large-scale street-view analysis using the gpt-5-mini API~\cite{openai_pricing}. Based on our measurements, processing one image requires on average 200 tokens, including both prompt and response. Assuming one street-view image every 20\,m, the cost of analyzing a road network of length $L$ is
\[
\text{Cost}(L)=\frac{L}{20\text{ m}}\times 200~\text{tokens}\times\frac{\$2}{10^6}.
\]
For a 100{,}000\,km network, this yields an estimated cost of roughly \$2{,}000. Note that this is a coarse, order-of-magnitude approximation; storage, retrieval, bandwidth, and other system-level costs are not included.

\input{sections/things/tab_cost}

Using publicly compiled national road-length statistics~\cite{wiki:road_network_size}, we approximate the \gls{vllm} processing cost for Poland, France, and Sweden. Table~\ref{tab:country_cost} reports the estimated cost assuming full-network coverage. In practice, an ADAS deployment analyzes only selected operational regions (e.g., major cities or specific roads), and street-view coverage is incomplete in some areas, so the actual cost is typically far lower than the full-coverage estimate. Although we do not quantify real-world test-drive costs, on-road data collection is generally considered far more expensive and logistically demanding than automated street-view analysis. Consequently, a processing cost of a few thousand dollars is typically acceptable in practice.



%% file: sections/things/fig_eval_countries.tex
\begin{figure}[htbp]
    \centering
    \includegraphics[width=\linewidth]{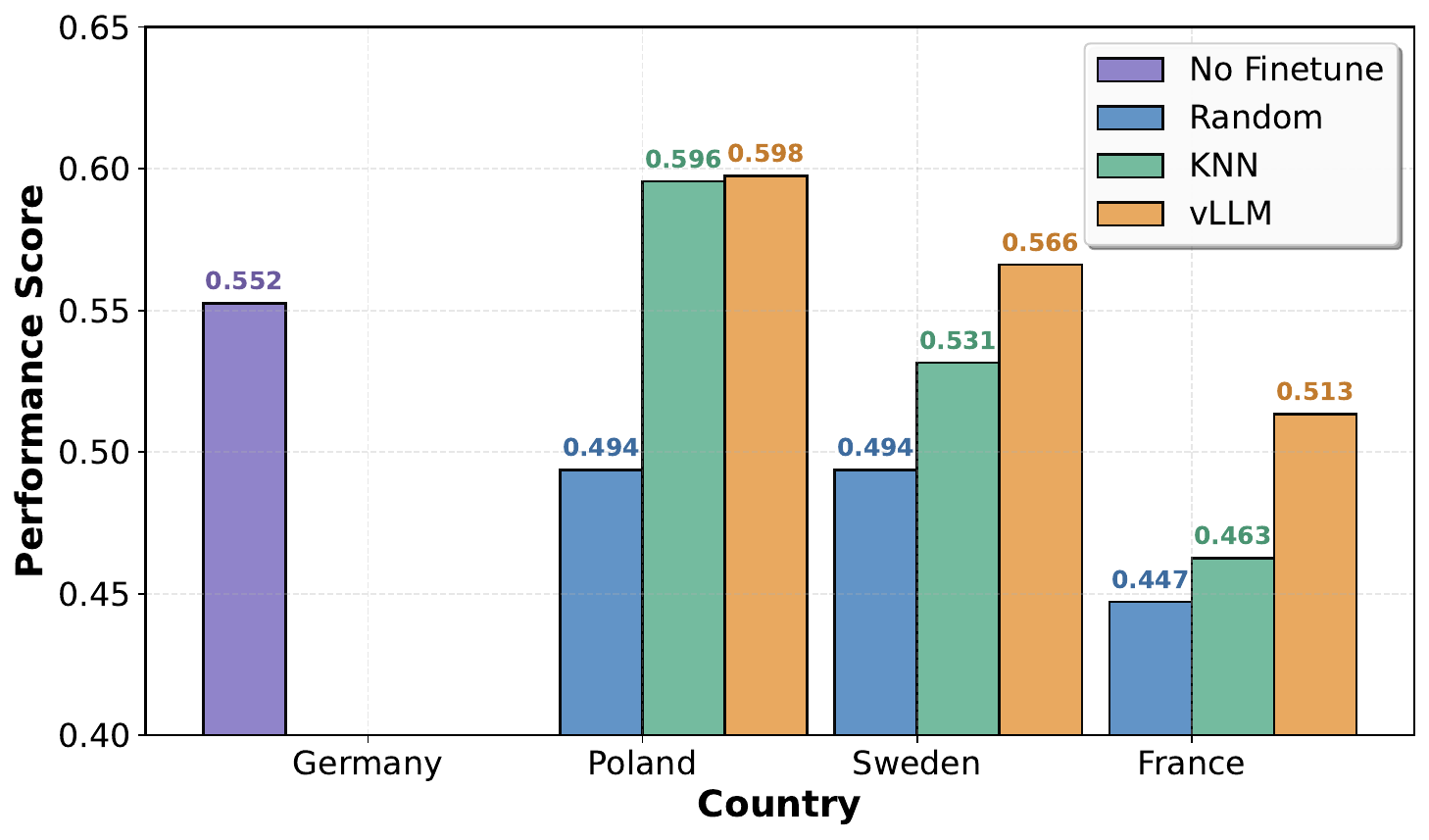}
    \caption{The \gls{tsd} model is pretrained on ZOD (Germany) dataset, and then finetuned with 25\% of ZOD (Poland, Sweden, France) dataset. }
    \label{fig:eval_countries}
\end{figure}

%% file: sections/things/fig_eval_exps_v2.tex
\begin{figure*}[t]
\centering
\scriptsize
\setlength{\tabcolsep}{3pt}

\begin{tabular}{ccccc}
\subcaptionbox{Detected sign\label{fig:eval_exp_01}}{%
  \includegraphics[width=0.18\linewidth]{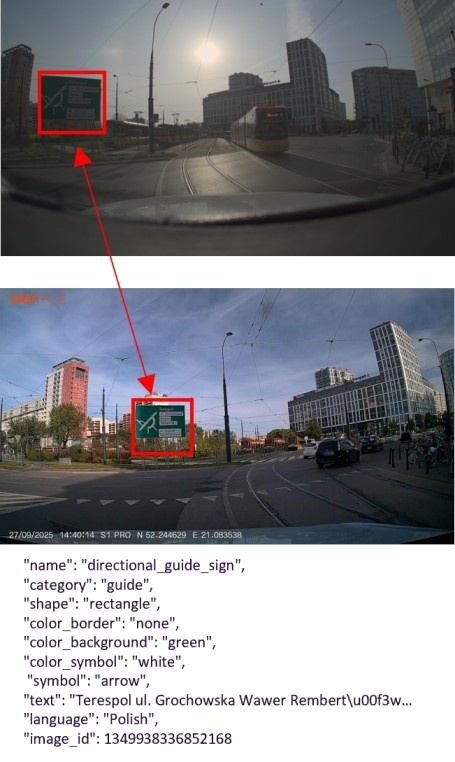}
} &
\subcaptionbox{Detected sign\label{fig:eval_exp_02}}{%
  \includegraphics[width=0.18\linewidth]{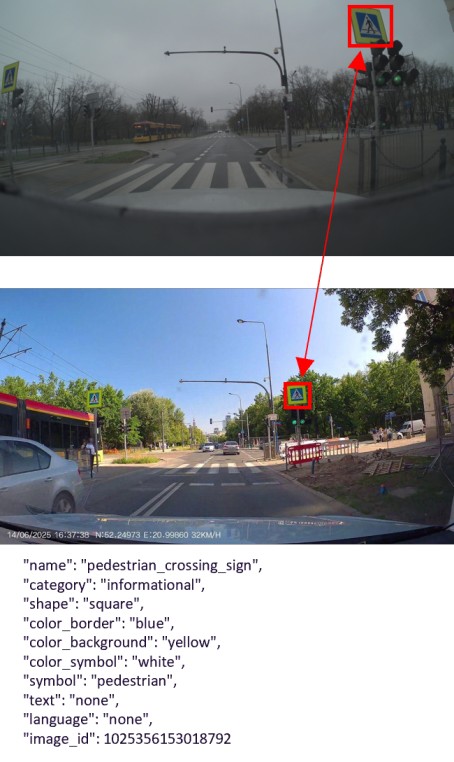}
} &
\subcaptionbox{Viewpoint mismatch\label{fig:eval_exp_03}}{%
  \includegraphics[width=0.18\linewidth]{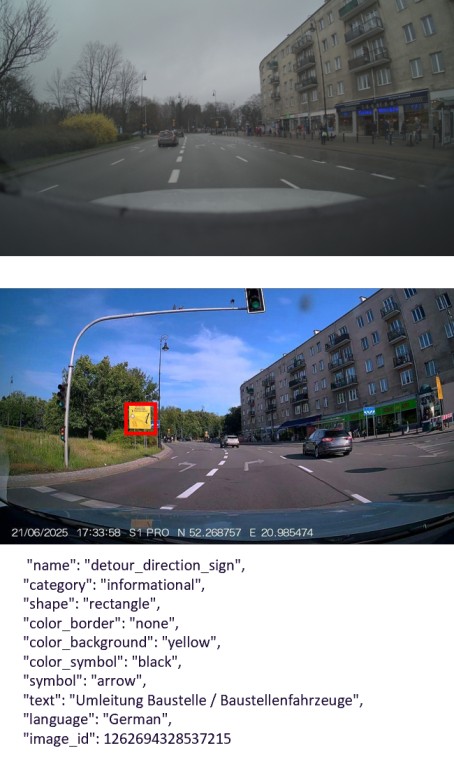}
} &
\subcaptionbox{Temporary removal\label{fig:eval_exp_04}}{%
  \includegraphics[width=0.18\linewidth]{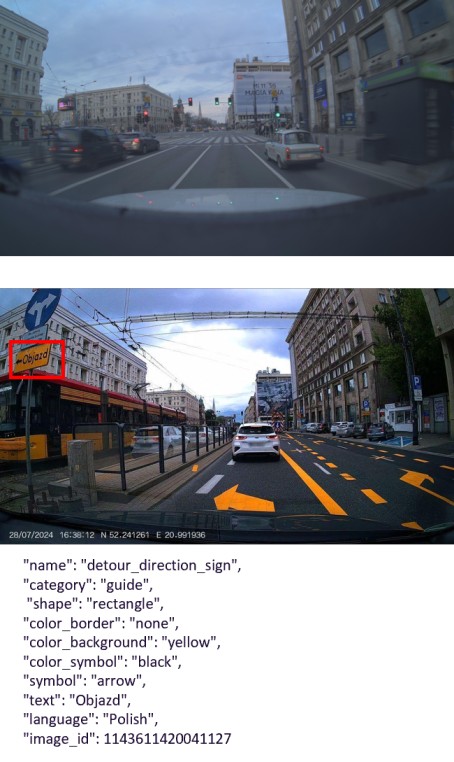}
} &
\subcaptionbox{KNN failure\label{fig:eval_exp_05}}{%
  \includegraphics[width=0.18\linewidth]{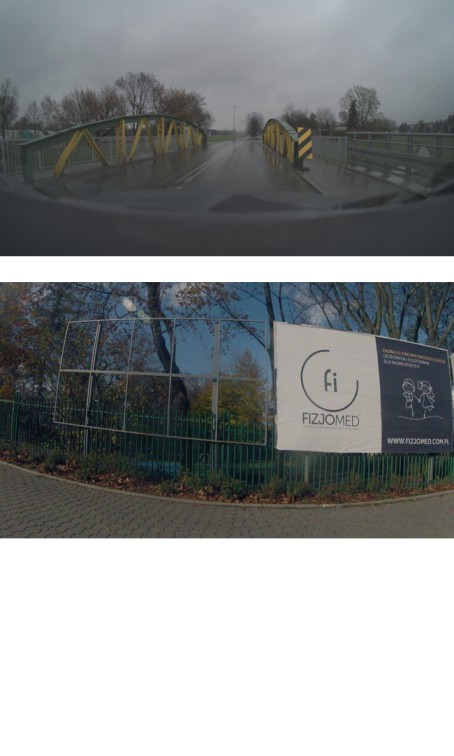}
}
\end{tabular}

\caption{Qualitative POI retrieval examples in Poland.
(a)--(d) Results from the proposed VLLM-based method; (e) KNN baseline.
Top: ZOD driving images. Bottom: Mapillary street-view.
(a) Correct guide sign.
(b) Pedestrian crossing with yellow background.
(c) Missing due to viewpoint mismatch.
(d) Missing due to temporary removal.
(e) High-score KNN false positive.}
\label{fig:eval_exps}
\end{figure*}

%% file: sections/things/tab_cost.tex
\begin{table}[htbp]
\centering
\caption{Estimated \gls{vllm} processing cost for full road network analysis (1 image per 20\,m, 200 tokens per image, \$2 per million tokens). National road length data from~\cite{wiki:road_network_size}.}
\label{tab:country_cost}
\begin{tabular}{lcc}
\toprule
\textbf{Country} & \textbf{Road Length (km)} & \textbf{Estimated Cost (USD)} \\
\midrule
Poland & 429{,}800 & \$8{,}596 \\
France & 1{,}053{,}215 & \$21{,}064 \\
Sweden & 573{,}134 & \$11{,}463 \\
\bottomrule
\end{tabular}
\end{table}

%% file: sections/6_conclusion.tex
\section{Conclusion}
\label{sec:conclusion}


This paper presented an efficient street-view-guided data acquisition strategy for \gls{adas} deployment. Instead of relying solely on costly large-scale test drives, we leverage publicly available street-view imagery to identify \glspl{poi} in the target country prior to real-world data collection. We designed a \emph{collect-detect} protocol and built a co-located driving and street-view dataset. The proposed \gls{vllm}-based method requires only half of the data to achieve the same level of model improvement.
Future work will extend this strategy to additional perception tasks and incorporate richer scene attributes for improved \gls{poi} discovery.